\documentclass[letterpaper, 10 pt, conference]{ieeeconf}  

\IEEEoverridecommandlockouts                              

\usepackage{graphics} 
\usepackage{epsfig} 
\usepackage{algorithm} 
\usepackage{cite}
\usepackage{algorithmicx}  
\usepackage{algpseudocode} 
\usepackage{times} 
\usepackage{amsmath} 
\usepackage{amssymb}  
\usepackage{hyperref}
\usepackage{url}
\usepackage{graphicx}
\usepackage{subfigure}
\usepackage{multicol}
\newtheorem{theorem}{Theorem}
\newtheorem{corollary}{Corollary}
\newtheorem{proposition}{Proposition}
\def\sA{{\mathbb{A}}}
\def\sS{{\mathbb{S}}}
\def\sH{{\mathbb{H}}}
\def\sP{{\mathbb{P}}}
\def\sO{{\mathbb{O}}}

\title{\LARGE \bf
Robust Reinforcement Learning under model misspecification
}

\author{Lebin Yu, Jian Wang and Xudong Zhang
\thanks{*All authors are with the Department of Electronic Engineering, Tsinghua University}
}

\begin{document}

\maketitle
\thispagestyle{empty}
\pagestyle{empty}

\begin{abstract}

Reinforcement learning has achieved remarkable performance in a wide range of tasks these days. Nevertheless, some unsolved problems limit its applications in real-world control. One of them is model misspecification, a situation where an agent is trained and deployed in environments with different transition dynamics. We propose an novel framework that utilize history trajectory and Partial Observable Markov Decision Process Modeling to deal with this dilemma. Additionally, we put forward an efficient adversarial attack method to assist robust training. Our experiments in four gym domains validate the effectiveness of our framework.  

\end{abstract}

\section{Introduction}
Over the past few years, Deep reinforcement learning has advanced at an unprecedented speed. Unlike traditional RL, DRL utilizes deep neural networks as function approximators. It has shown reliable performance in many tasks such as game playing \cite{mnih2013playing,silver2016mastering,vinyals2019grandmaster} as well as real-world control \cite{gu2017deep,finn2017deep}.  

Despite the brilliant achievements DRL has made in simulation environments, there remains numerous challenges when applying it to real-world tasks \cite{dulac2019challenges}. One major challenge is model misspecification. An agent is usually trained in an ideal virtual environment where the dynamics are consistent and collecting training samples is convenient. However, the real environment where the agent is deployed cannot be the same with experimental ones: the dynamics are possibly different. As an example, the mass of a self-nav Unmanned Ground Vehicle (UGV) may vary due to the cargo it carries. Additionally, terrible road conditions could change ground friction and affect steering performance. These factors could severely degrade an agent's performance \cite{peng2018sim} and have become roadblocks to the application of RL in real-world control. Such a problem, an agent is trained in an ideal environment but performs in another environment whose parameters are different, is referred to as model misspecification. 

Several recent works focus on training robust RL agents against model misspecification \cite{mankowitz2019robust,roy2017reinforcement}. Most of them follow the concept of maximizing rewards in the worst condition. This is perceived to be a reasonable risk-averse criterion. Nevertheless, we notice that agents can learn about the unknown environment parameters from history trajectory and choose a more specific strategy. Consequently, we come up with a robust RL framework that utilizes history trajectory. In particular, our principal contributions are four-fold:

\begin{itemize}
    \item We put forward a novel adversarial attack method against environment parameters.
    \item We confirm that the traditional robust reinforcement learning framework provides sub-optimal strategies, while treating environment parameters as a part of the system state brings better performance.
    \item We are the first to introduce Partial Observable Markov Decision Process (POMDP) and utilize history trajectory to deal with model misspecification. We also extend an acclaimed RL method Soft Actor-Critic (SAC) for POMDP to yield Recurrent SAC (RSAC). Additionally, we design a specified neural network structure to learn from history.
    \item We conduct experiments in multiple environments and demonstrate the effectiveness of the proposed method. We also make several investigative experiments to better understand our proposed framework.
\end{itemize}

\section{Related Work}
Robust Reinforcement learning has received increasing attention these years. Among relative researches, there are several kinds of robustness. A common definition is the robustness against random noises and attackers \cite{pan2019risk,mandlekar2017adversarially,tessler2019action} . Some define it as reducing rewards variance \cite{cheng2019control,smirnova2019distributionally}. Some others investigate robust RL with model misspecification \cite{mankowitz2019robust,roy2017reinforcement}. For the remainder of this paper, when we mention robustness, if not specified, we refer to the last kind of robustness.

Adversarial attack is a common trick often utilized in robust RL. Numerous works have confirmed its efficiency \cite{havens2018online,huang2017adversarial}. Some choose to disturb the observation \cite{pattanaik2018robust}, while some others adversarially manipulate agents' output action \cite{gu2019adversary,pinto2017robust}. Several robust training methods go without adversarial attack, though. Following the perception of max-min rewards, \cite{mankowitz2018learning} pre-define an uncertainty set of environment parameters, and agents are required to get the highest rewards with the worst parameters at each transition. \cite{mankowitz2019robust} continues the method. In this paper, we combine these two frameworks to an innovative attack technique that directly modify environment parameters.  

We would like to elucidate the difference between our work and meta-reinforcement learning. \cite{nagabandi2018learning} have also paid attention to the situation where an agent is trained and tested in different environments. They utilize meta-training so that agents can quickly adapt to new settings with only a few training steps. The main difference is that their model misspecification is much more severe than ours, e.g. the change of action dimensions. In such a dilemma, robust training is not adequate for agents to make valid decisions while additional training is required. However, when the environment changes are not that remarkable, which is often the case, our method is more efficient.  

There are a few previous papers using history trajectory in reinforcement learning. \cite{heess2015memory} put forward Recurrent Deterministic Policy Gradient algorithm (RDPG) to deal with POMDP. Nevertheless, it is an offline approach with slow convergence. Its improved version, Fast-RDPG is proposed by \cite{wang2019autonomous}. The authors apply it to a self-navigation task in a complex unknown urban environment and get promising results. Inspired by them, we choose to utilize history trajectory to handle model misspecification. To the best of authors' knowledge, no one has done this before.   

\section{Preliminaries}
In this section, we present some fundamental models and methods that are crucial to our research. We utilize Markov Decision Process (MDP) and Partial Observable Markov Decision Process (POMDP) for modelling, and an acclaimed RL framework, Soft Actor-Critic\cite{haarnoja2018soft}, as our basic algorithm. Additionally, we introduce the traditional robust RL framework Robust MDP (RMDP) and a corresponding algorithm which will be tested in our experiments. 
\subsection{MDP and POMDP}
A Markov Decision Process (MDP) is composed of a tuple $\{ \sS, \sA, P, R \} $ where $\sS$ is the state set, $\sA$ is the action set, $p(s_{t+1}|s_t, a_t)$ is the transition probability distribution and $r(s_t, a_t)$ is the reward function. This is the fundamental model of RL.

The solution of MDP is an optimal policy $\pi$. The policy could be either deterministic or stochastic. A deterministic policy will choose a certain action based on the state: $a = \pi(s)$, while a stochastic policy will output a distribution: $a \sim \pi(\cdot|s)$. Multiple pieces of research have revealed that the latter one has better performance. Therefore, this paper primarily focuses on stochastic policy.

Standard RL finds the optimal policy by maximizing the objective function of a policy $\pi$:
\begin{equation}
    J(\pi) =\mathbb{E}_{(s_t, a_t)\sim \rho_{\pi} }  \biggl[ \sum^{\infty}_{t=0}\gamma^t r(s_t, a_t) \biggr]
\end{equation}
Where $\gamma$ is the discount factor. During training, value functions and action-value functions are used for estimating the values of states and actions
\begin{equation}
    V^{\pi}(s_t) = \mathbb{E} \biggl[ \sum^{\infty}_{l=0}\gamma^t r(s_{t+l}, a_{t+l}) \bigg| s_t \biggr]
\end{equation}
\begin{equation}
    Q^{\pi}(s_t, a_t) = \mathbb{E} \biggl[ \sum^{\infty}_{l=0}\gamma^t r(s_{t+l}, a_{t+l}) \bigg| s_t, a_t \biggr]
\end{equation}
and Value function is the expectation of Q-function: 
\begin{equation}
    V^{\pi}(s_t) = \mathbb{E}_{a_t \sim \pi} [ Q^{\pi}(s_t, a_t) ]
\end{equation}

Partial Observable Markov Decision Process (POMDP) \cite{astrom1965optimal} is a generalization of MDP where the state $s_t$ of the system cannot be fully observed by the agent. Instead, it receives an observation $o_t$  with a distribution $p(o_t|s_t)$. Correspondingly, the goal of RL in POMDP settings is to find optimal policy that output actions based on observations: $a_t \sim \pi(\cdot|o_t)$. Robust RL is generally modeled as MDPs, though, if we treat the misspecified environment parameters as a part of the state, it becomes POMDP for we cannot access the exact values of them. We will illustrate the reasons for this modeling in Section~\ref{Why_use_POMDP}

\subsection{Soft Actor-Critic}
SAC, proposed by \cite{haarnoja2018soft}, is one of the most efficient model-free RL methods. It follows the idea of maximum entropy RL \cite{ziebart2010modeling} which augments the objective function with an entropy term:
\begin{equation}
    J(\pi) = \mathbb{E}_{(s_t, a_t)\sim \rho_{\pi} } \biggl[ \sum^{\infty}_{t=0}\gamma^t r(s_t, a_t) + \alpha H(\pi(\cdot|s_t)) \biggr]
\end{equation}
where $\alpha$ is the temperature parameter determining the relative importance if the entropy term against the reward. Then the soft value function $V^\pi(s_t)$ is redefined as:
\begin{equation}
\label{rawV}
    V^\pi(s_t) = \mathbb{E}_{a_t \sim \pi} [ Q^{\pi}(s_t, a_t) - log\pi(a_t|s_t)]
\end{equation}
and the soft Q-function $Q^\pi(s_t, a_t)$ can be calculated using Bellman backup equation:
\begin{equation}
\label{bellman}
    \mathcal{T}^\pi Q^\pi(s_t, a_t) = r(s_t, a_t) + \gamma \mathbb{E}_{s_{t+1} \sim p}[V(s_{t+1})]
\end{equation}
Then the maximum entropy policy should meet the undermentioned condition:
\begin{equation}
    \pi(\cdot|s_t) \propto Q_\pi(s_t, \cdot)
\end{equation}
The proof is provided in \cite{haarnoja2017reinforcement}. Consequently, the policy can be updated by minimizing KL divergence to the exponential of the Q-function:
\begin{equation}
    \pi_{new} = \arg \mathop{\min}_{\pi' \in \Pi} D_{\mathrm{KL}} \biggl( \pi'(\cdot|s_t) \bigg| \bigg| \frac{exp(Q^{\pi_{old}}(s_t, \cdot))}{Z^{\pi_{old}}(s_t)} \biggr)
\end{equation}
The authors adopts several practical tricks techniques for the purpose of  accelerating convergence. They include a separate value function network even though there is no need in principle. Additionally, they make use of two Q-function networks. They are trained independently and the minor output will be used in the updating steps.  

We choose SAC as our base method for two reasons. Firstly, it manifests prime performance in a wide range of tasks. Secondly, its stochastic policies can provide baseline robustness. Therefore, comparing with SAC can convincingly illustrate the effectiveness of robust training methods. 

\subsection{Robust Markov Decision Process}
Robust Markov Decision Process (RMDP) is proposed by \cite{iyengar2005robust}. It is a variant of MDP where the transition probability is chosen from an uncertainty set $p \in \sP(s, a)$ in place of a consistent one. The goal of a Robust MDP is learning a robust policy that maximizes the worst-case performance. The robust value function is defined as follows:
\begin{equation}
\label{RMDP}
    V^\pi(s_t) = r(s_t,\pi(s_t)) + \gamma inf_{p\in \sP} \mathbb{E} [V^\pi(s_{t+1})|s_t,\pi(s_t)]
\end{equation}
This framework has been widely used in all kinds of robust reinforcement learning tasks \cite{pan2019risk,gu2019adversary,havens2018online} and shown promising results. Multiple works on model misspecification also follow this modeling \cite{mankowitz2019robust,mankowitz2018learning,pattanaik2018robust}. 

\subsection{An RMDP-based training method}
\label{SAC-AO}
\cite{pattanaik2018robust} put forward a robust training framework based on RMDP. The authors indicate that the lower bound in \ref{RMDP} can be obtained by adversarially changing observations. The worst case observation $s_{adv}$ is defined as the observation that causes the agent to choose the worst action: 
\begin{equation}
    s_{adv} = \mathop{\arg\min}_{s'} Q(s,\pi(s'))
\end{equation}
where $s$ is the real state, $Q(\cdot,\cdot)$ is the critic and $\pi(\cdot)$ is the actor. The optimal adversarial state $s_{adv}$ can be achieved by gradient descent:
\begin{equation}
       \nabla_s Q(s,a^*) = \frac{\partial Q}{\partial s} + \frac{\partial Q}{\partial a^*} \frac{\partial a^*}{\partial s}
\end{equation}
 The robust training process can be summarized as follows. Initially, the agent is trained without perturbations using ordinary RL methods. When it learns preliminary strategies, it has to make decisions and be trained with adversarially changed observations. The RL algorithms used in the original paper are DQN and DDPG. For ease of comparison, we re-implement it based on SAC. We will refer to it as SAC-AO in our experiments. Additional details are described in Algorithm~\ref{alg:atkobs}. Besides, in the original implementation, a beta distribution $beta(\alpha, \beta) $ is used for sampling while the exact value of $\beta$ is not given. Consequently, we use a uniform distribution $U(0,\beta)$ instead.   
 
\begin{algorithm}[tbhp]
\caption{SAC-AO}
\label{alg:atkobs}
\begin{algorithmic}
\Require training steps $t_{m1}, t_{m2}$, sample times $n$, maximum attack strength $\beta$
\For{$t=1:t_{m1}$}
\State train without perturbations
\EndFor
\State $s_0=ENV.reset()$
\For{$t=1:t_{m2}$}
\State Generate adversarial observations:
\State $a^* = \pi(s_{t}), Q_{min} = \infty$
\State $grad = \nabla_s Q(s_{t},a^*)/||\nabla_s Q(s_{t},a^*) ||$
\For{i = 1:n}
\State $l_i \sim U(0, \beta)$
\State $s^i_{adv}  = s_{t} - l_i \cdot grad $
\State $Q_i = Q(s_{t}, \pi(s^i_{adv}))$
\If{$Q_i < Q_{min}$}
\State $Q_{min} = Q_i$, $s_{adv} = s^i_{adv} $
\EndIf
\EndFor
\State begin robust training:
\State $a_t \sim \pi(\cdot|s_{adv})$
\State $s_{t+1},r_t = ENV.step(a_t)$
\State update networks
\EndFor
\end{algorithmic}
\end{algorithm}

\section{Methods}
\label{Why_use_POMDP}
In this section, we give our reasons for using history trajectory and POMDP modeling rather than RMDP. Firstly, we substantiate that treating the environment parameter $\omega$ as a part of the state is promised to obtain a better policy than RMDP. Specifically, the optimum hyper policy $\pi^*_h(\cdot|s_t)$ that performs the best under all conditions can be obtained in theory. Secondly, we point out that $\omega$ is unobservable and hence do POMDP modeling. Additionally, we extend Soft Actor-Critic to POMDP scenarios and yield Recurrent SAC (R-SAC). Note that all derivations in this section are within the maximum entropy framework.
\subsection{Optimal Strategy under Model Misspecification}
We denote the misspecified environment parameters as $\omega$, the corresponding environment as $Env_\omega$. Take the reality into account, we have two assumptions for $\omega$: (1)$\omega$ does not vary frequently. (2)The new value of $\omega$ solely relies on the present value, i.e.
$H(\omega_{t+1}|\omega_{t})=H(\omega_{t+1}|\omega_{t},\omega_{t-1})$. Define $o_t$ as the ordinary observation of the system, then we have the following proposition:
\begin{proposition}
\label{MarkovA}
$\{ \sS=\sO \times \Omega, \sA, P, R\}$ is an MDP.
\end{proposition}
We suppose a hyper agent can access the complete state of the system $s_t = [o_t, \omega_t]$ and has a corresponding hyper policy $\pi_h(\cdot|s_t)$. Then we can define optimal hyper policy $\pi^*_h(\cdot|s_t)$ using soft Q-function specified in \ref{rawV} and \ref{bellman}:
\begin{equation}
\label{HyperBest}
    Q^{\pi^*_h}([o, \omega], a) \ge Q^{\pi_h}([o, \omega], a) \quad \forall \pi_h, o, a, \omega
\end{equation}
We use $Q_\omega$ to denote the action-value function of $Env_\omega$ and $\pi^{\omega}_h$ to denote the projection of hyper policy $\pi_h$ to $Env_\omega$: $\pi_h^{\omega}(\cdot|o_t)=\pi_h(\cdot|o_t,\omega) $
It can be derived from the above equation that $\pi^{\omega*}_h$, the projection of the best hyper policy $\pi^*_h$ to $Env_\omega$, is also the best policy for $Env_\omega$:
\begin{equation}
    Q^{\pi^{\omega*}_h}_\omega(o,a) \ge Q^\pi_\omega(o, a) \quad \forall \pi, o, a 
\end{equation}
In this setting, a major drawback of RMDP appears: it provides sub-optimal policies. Since it  make decisions based on the present observation, it cannot react to changes of environment dynamics. It only promises to perform the best action in the worst condition, not the best action in all settings. In contrast, treating the parameters as a part of the state leads to the best hyper policy that suits all conditions. Moreover, the best policy is obtainable. To prove it, we cite a theorem in \cite{haarnoja2018soft} here:

\begin{theorem}
(Soft Policy Iteration) Repeated application of soft policy evaluation and soft policy improvement from any $\pi \in \Pi$ converges to a policy $\pi^*$ such that $Q^{\pi^*}(s,a) \ge Q^{\pi}(s, a) $ for all $\pi \in \Pi$ and $(s,a) \in \sS \times \sA $, assuming $|\sA| < \infty $
\end{theorem}

Simply replace $s$ with $[o,\omega]$ and we get the following corollary:

\begin{corollary}
Repeated application of soft policy evaluation and soft policy improvement from any $\pi \in \Pi$ converges to a policy $\pi^*$ such that $Q^{\pi^*_h}([o, \omega], a) \ge Q^{\pi_h}([o, \omega], a) $ for all $\pi \in \Pi$ and $(s,a) \in \sS \times \sA $, assuming $|\sA| < \infty $
\end{corollary}

\subsection{POMDP Modeling}
In reality, we cannot get the precise value of $\omega$ in an environment with model misspecification. Instead, we receive an observation $o_t$ having a distribution $p(o_t|s_t)$. Note that  history trajectory $h_t = [...,o_{t-1},a_{t-1},o_t]$ contains more information about $\omega$ than $o_t$, i.e. $I(\omega; h_t) \gg I(\omega; o_t)$, hence we should choose it as our network input. 

\subsection{Soft Actor-Critic for POMDP}
In this subsection we extend SAC for POMDP scenarios to get R-SAC. We can begin our derivation from $s_t = (h_t,\omega)$:  
\begin{proposition}
\label{MarkovB}
$\{ \sS=\sH \times \Omega, \sA, P, R\}$ is an MDP.
\end{proposition}
Define the soft action-value function as: $Q^\pi(h_t,a_t) = E_{\omega \sim h_t}[Q(h_t, \omega, a_t)]$, the value function as:
\begin{equation}
\label{VFdefine}
    \begin{split}
        V^\pi(h_t)& = E_{\omega \sim h_t}[V^\pi(h_t, \omega)]    \\
        & = E_{\omega \sim h_t}[E_{a_t \sim \pi}[Q(h_t,\omega,a_t)-log\pi(a_t|h_t)] ] \\
        & = E_{a_t \sim \pi}[E_{\omega \sim h_t}[Q(h_t,\omega,a_t)-log\pi(a_t|h_t)  ]] \\
        & = E_{a_t \sim \pi}[Q(h_t,a_t)-log\pi(a_t|h_t)]
    \end{split}
\end{equation}
This equation has the same form as \ref{rawV}. Then we can get the corresponding bellman backup equation:
\begin{equation}
\label{Newbellman}
\begin{split}
  \mathcal{T}^\pi Q^\pi(h_t, a_t) & = E_{\omega \sim h_t}[   \mathcal{T}^\pi Q^\pi(h_t,\omega,a_t)] \\
  & =  E_{\omega \sim h_t}[r(h_t,\omega,a_t)+\gamma E_{h_{t+1}}[V^\pi(h_{t+1},\omega)]] \\
    & = r(h_t,a_t)+\gamma E_{h_{t+1}}[V^\pi(h_{t+1})]
\end{split}
\end{equation}
We consider a parameterized value function $V_\psi(h_t)$, Q-function $Q_\theta(h_t,a_t)$, policy $\pi_\phi(a_t|h_t)$ with parameters $\psi$, $\theta$ and $\phi$. We also use two Q networks and a target value network $V_{\hat{\psi}}(h_t)$ to accelerate training. Since the above equation has the same form as \ref{bellman}. In this way, following the derivations of \cite{haarnoja2018soft}, we can obtain the gradients of our networks: $\nabla_\psi J_V(\psi)$, $\nabla_\theta J_Q(\theta)$ and $\nabla_\phi J_\pi(\phi)$. 
\begin{equation}
    \begin{split}
        \nabla_\psi J_V(\psi)  =& \nabla_\psi V_\psi(h_t)V_\psi(h_t) \\ 
        &-\nabla_\psi V_\psi(h_t)(Q_\theta(h_t,a_t)-log\pi_\phi(a_t|h_t)) \\
        \nabla_\theta J_Q(\theta)  =& \nabla_\theta Q_\theta(h_t,a_t)Q_\theta(h_t,a_t) \\
        &- \nabla_\theta Q_\theta(h_t,a_t)(
        r(h_t,a_t)+\gamma V_{\hat{\psi}}(h_{t+1})) \\
        \nabla_\phi J_\pi(\phi) =& \nabla_\phi log \pi_\phi(a_t|h_t)\\ &+\nabla_\phi a_t(\nabla_{a_t}log\pi_\phi(a_t|h_t)-\nabla_{a_t}Q(h_t,a_t))
    \end{split}
\end{equation}

\section{Adversarial Training}
In this section, we explain our adversarial attack and robust training method in detail.  
\paragraph{Attack Target} Our adversary has access to changing environment parameters, e.g. the mass of the agent. Its purpose is to make the new environment different from the old one as much as possible from the agent's view. Denote the current parameter as $\omega$, the previous parameter before last attack as $\omega_{-1}$, the modified one as $\omega'$, current observation and agent's action as $o_t, a_t$, the next observation under $\omega$ and $\omega'$ as $o_{t+1}$ and $o'_{t+1}$. The adversary chooses $\omega'$ by maximizing the objective function $J(\omega')$: 
\begin{equation}
    \omega'_{opt} = \arg \max_{\omega'}{J(\omega'), \omega' \in \Omega}
\end{equation}
where $\Omega$ is a pre-defined parameter set and $J(\omega') = || o_{t+1} - o'_{t+1} || $. During testing, we find this may result in parameters changing between two points on the boundary of $ \Omega $. An exploration term is added to prevent such an occurrence: $\varepsilon = || \omega' - \omega_{-1} ||$. So the final objective function becomes:
\begin{equation}
    J(\omega') = (1-\lambda) || o_{t+1} - o'_{t+1} || + \lambda\varepsilon
\end{equation}
where $\lambda$ is a weighting factor. 

In practice, how to obtain $\omega'$ depends heavily on our access to the training environment. In an environment where the analytical expression of the transition model is known, we can calculate $\nabla_{\omega'} J(\omega')$ and use Projected Gradient Descent (PGD) method. This is named  as \textit{gradient-based attack}.  In an environment where $o'_{t+1}$ can be obtained without changing the system state, which is often the case, we can sample several $\omega'$ and find the best one. This is named as \textit{sample-based attack}. The pseudo-code is given in Algorithm~ \ref{alg:atkenv}. In an environment with minimum authority where $o'_{t+1}$ is not obtainable, we can re-define $J(\omega')$:
\begin{equation}
    J(\omega') = (1-\lambda) || \omega' - \omega || + \lambda || \omega' - \omega_{-1} ||
\end{equation}
and use sampling to obtain $\omega'$. This is named as \textit{black-box attack}.

\begin{algorithm}[tbhp]
\caption{Attack Environment dynamics}
\label{alg:atkenv}
\begin{algorithmic}
\Function{Env-Attack}{$o_t, a_t, n, \omega_{-1}, \Omega$}
\State set $J^* = 0, \omega^* = 0$
\For{i = 1: n}
\State sample $\omega' \in \Omega$
\State obtain $ o_{t+1} \sim p(o_t, a_t | \omega') $
\State $J(\omega') = (1-\lambda) || o_{t+1} - o'_{t+1} || + \lambda || \omega' - \omega_{-1} || $
\If{$J(\omega')> J^*$}
\State $J^* = J(\omega'), \quad \omega^* = \omega' $
\EndIf
\EndFor
\State \textbf{return} $\omega^*$
\EndFunction
\end{algorithmic}
\end{algorithm}

\paragraph{Attack Timing}
Since we use history trajectory to estimate $\omega$ and make decisions, the environment cannot change frequently. Instead of attacking periodically, we prefer a more intelligent option. \cite{kos2017delving} ascertain that attacking at specific times is much more efficient. The key innovation is attacking when the agent is certain about what to do. For deterministic policy (e.g. DQN), this can be measured by the range of Q: $\mathop{\max}_a{Q(s_t, a)}-\mathop{\min}_a{Q(s_t, a)}$. For stochastic policy (e.g. SAC), we can refer to the standard deviation $\sigma$ of the output action. Following their idea, we modify environment parameters when $\sigma$ is smaller than a threshold $\sigma_{th}$. It has another benefit: the attack will not happen at the beginning of the training. Only when the agent learns elementary policies and outputs actions with low std will the adversary begins interfering. To avoid attacking too frequently, we also set a minimum attack interval $t_{th}$.

The complete algorithm is given in Algorithm~\ref{alg:total-train}. 

\begin{algorithm}[tbhp]
\caption{R-SAC with adversarial attack}
\label{alg:total-train}

\begin{algorithmic}
\Require threshold $\sigma_{th}$, attack minimum interval $t_{th}$, parameter set $\Omega$, training steps $t_m$, max memory length $l_{max}$, attack sample times $n$
\State $t_l = 0$, $o_0 = ENV_\omega.reset()$, $h_0=[o_0]$ 
\For{$t = 0: t_m$}
\State $[a_t, \sigma_t ] \sim \pi_\phi(h_t)$ 
\If{ $t - t_l > t_{th} $ and $\sigma_{th}$ } 
\State $\omega$ = \Call{Env-Attack}{$o_t, a_t, n, \omega_{-1}, \Omega$}
\State update $t_l, \omega_{-1}$
\EndIf
\State $o_{t+1}, r_t = ENV_\omega.step(a_t)$
\State $h_{t+1} = [h_t,a_t,o_{t+1}] $
\If{$len(h_t)>l_{max}$}
\State delete $a_{t-l_{max}}$, $o_{t-1-l_{max}} $ in $h_t$
\EndIf 
\State store transition $[h_t, a_t, h_{t+1}, r_t]$ into replay buffer $R$
\If{train}
\State sample a batch of transitions from $R$
\State $\psi = \psi - \lambda_V \nabla_\psi J_V(\psi)$
\State $\theta_i = \theta_i - \lambda_Q \nabla_{\theta_i} J_Q(\theta)$ for $i \in [1,2]$
\State $\phi = \phi - \lambda_\pi \nabla_\phi J_\pi(\phi)$
\State $\hat{\psi} = \tau\psi+(1-\tau)\hat{\psi} $
\EndIf
\EndFor
\end{algorithmic}

\end{algorithm}

\begin{figure}[tbhp]
  \centering
  \includegraphics[width=0.48\textwidth]{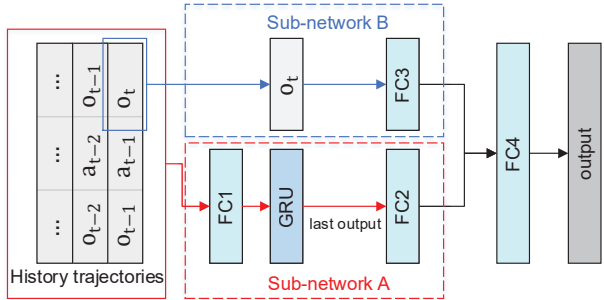} 
    \caption{The NN architecture for V network and $\pi$ network. For Q network, replace $o_t$ before FC3 with $[o_t,a_t]$. Adjacent layers are fully connected. Note that only the last-timestep output of GRU layer is used. }
    \label{NN-archi} 
\end{figure}

Additionally, we propose a specialized neural network architecture for R-SAC as shown in Fig.~\ref{NN-archi}. Sub-network A is designed to extract information of environment dynamics from history trajectory, while Sub-network B extracts features from observations. Note that removing Sub-network A leads to the raw network of SAC. Our experiments illuminate that such a combination can efficiently handle model misspecification. 

\section{Experiments}
\subsection{Main experiments}
The effectiveness of our proposed algorithms is tested in four gym \cite{1606.01540} domains: Pendulum, Cart-Pole, Hopper, and Walker. For ease of comparison, we change only one parameter of the environment during testing. Suppose the default environment parameter is $\omega_0$, and interference strength can be measured by the multiplier $\omega' / \omega_0$.

We compare the performance of four algorithms: (a) RSAC-AE, our proposed robust training framework with history trajectory and RNN. (b) SAC-AE, raw SAC architecture with our robust training method. It shares the same adversarial training with RSAC-AE but makes decisions based solely on current observations. (c) SAC-AO, raw SAC architecture with robust training by adversarially changing observations. The details can be found in Section~\ref{SAC-AO}. (d) SAC, the baseline. Our implementation for SAC is based on \cite{stable-baselines}. Our codes are available on https://github.com/PaladinEE15/RSAC.

\begin{figure}[t]
\centering
\begin{minipage}[t]{0.235\textwidth}
\centering
\includegraphics[width=1\textwidth,height=3cm]{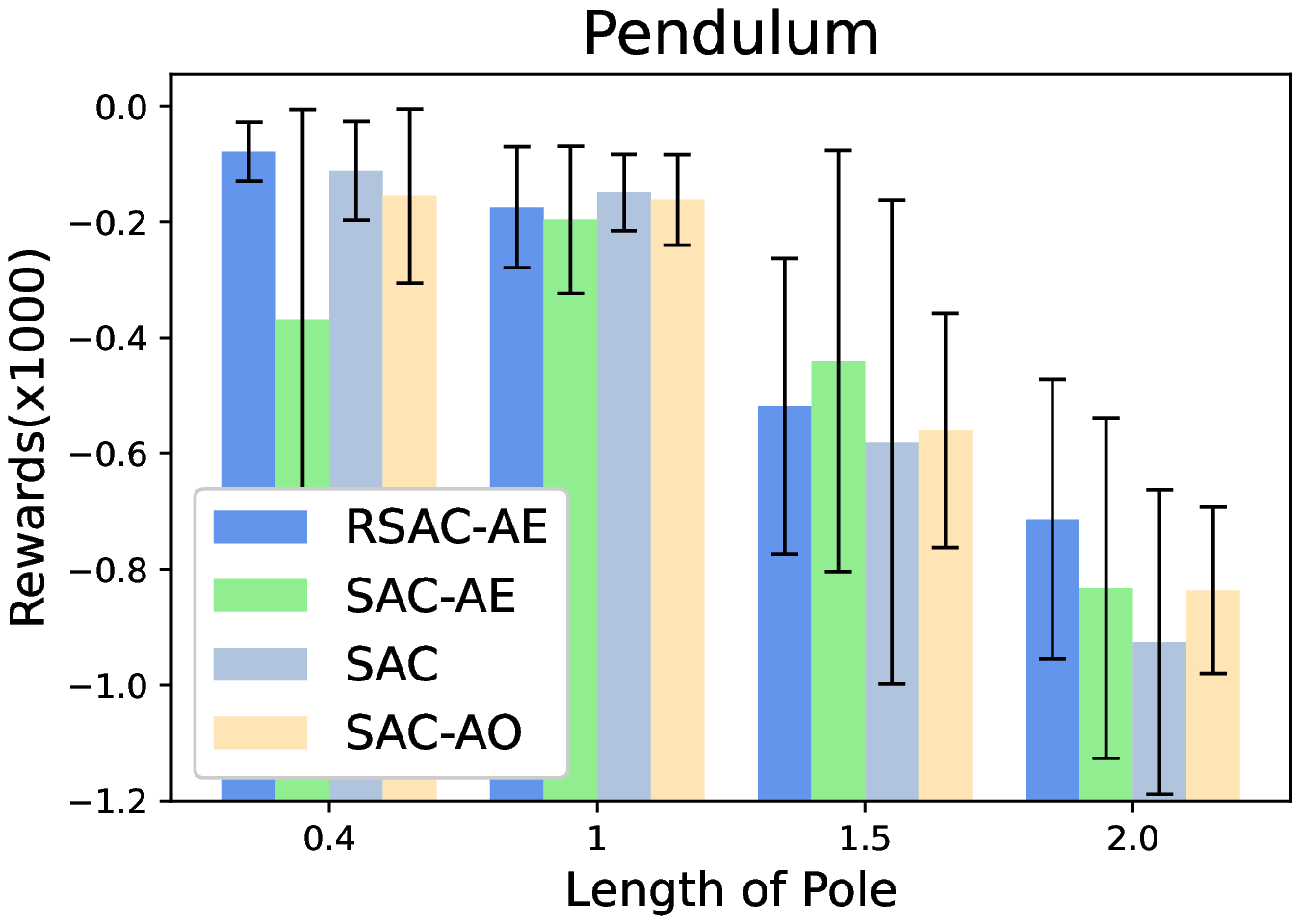}
\end{minipage}
\begin{minipage}[t]{0.235\textwidth}
\centering
\includegraphics[width=1\textwidth,height=3cm]{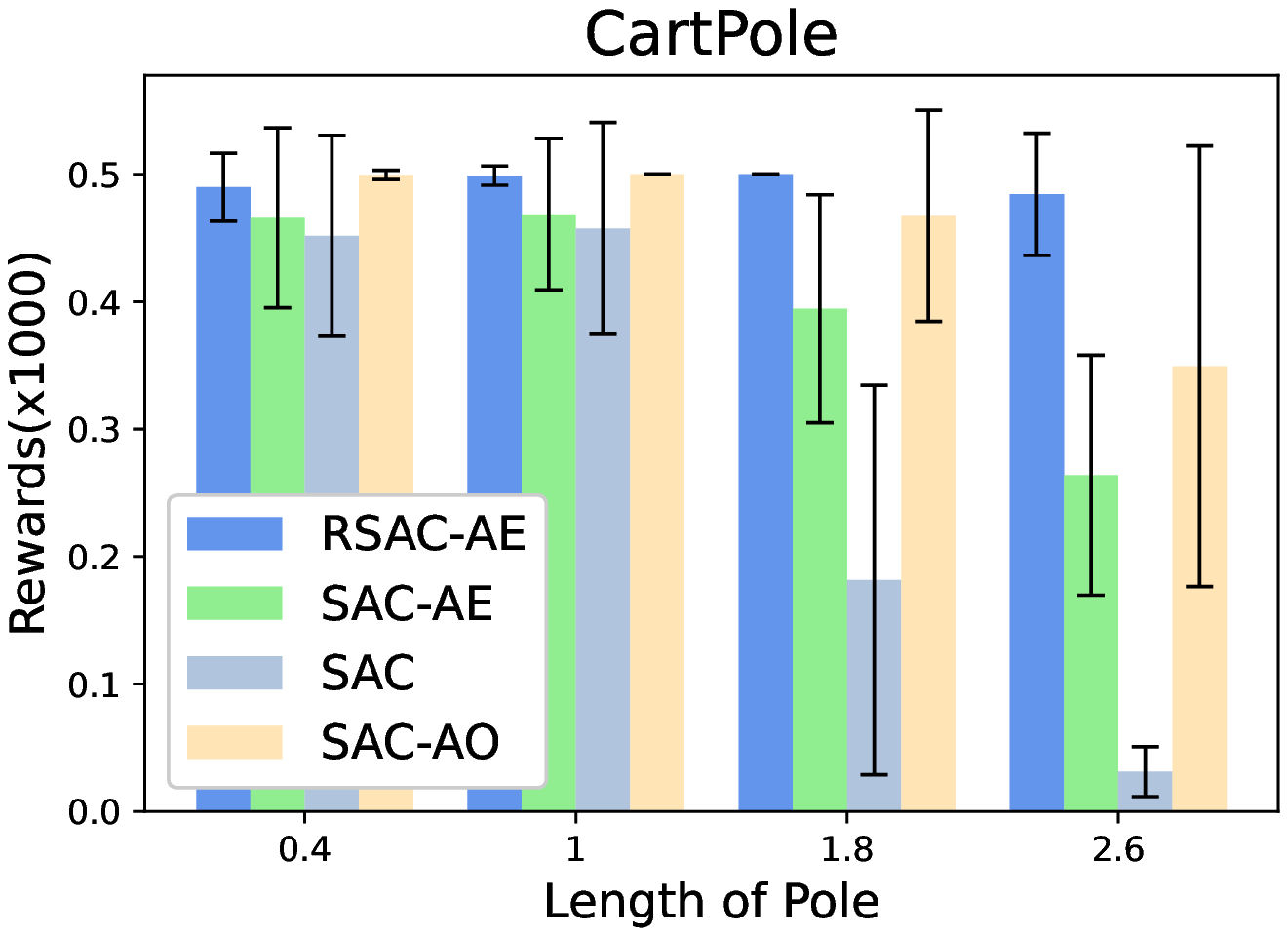}
\end{minipage}

\begin{minipage}[t]{0.235\textwidth}
\centering
\includegraphics[width=1\textwidth,height=3.3cm]{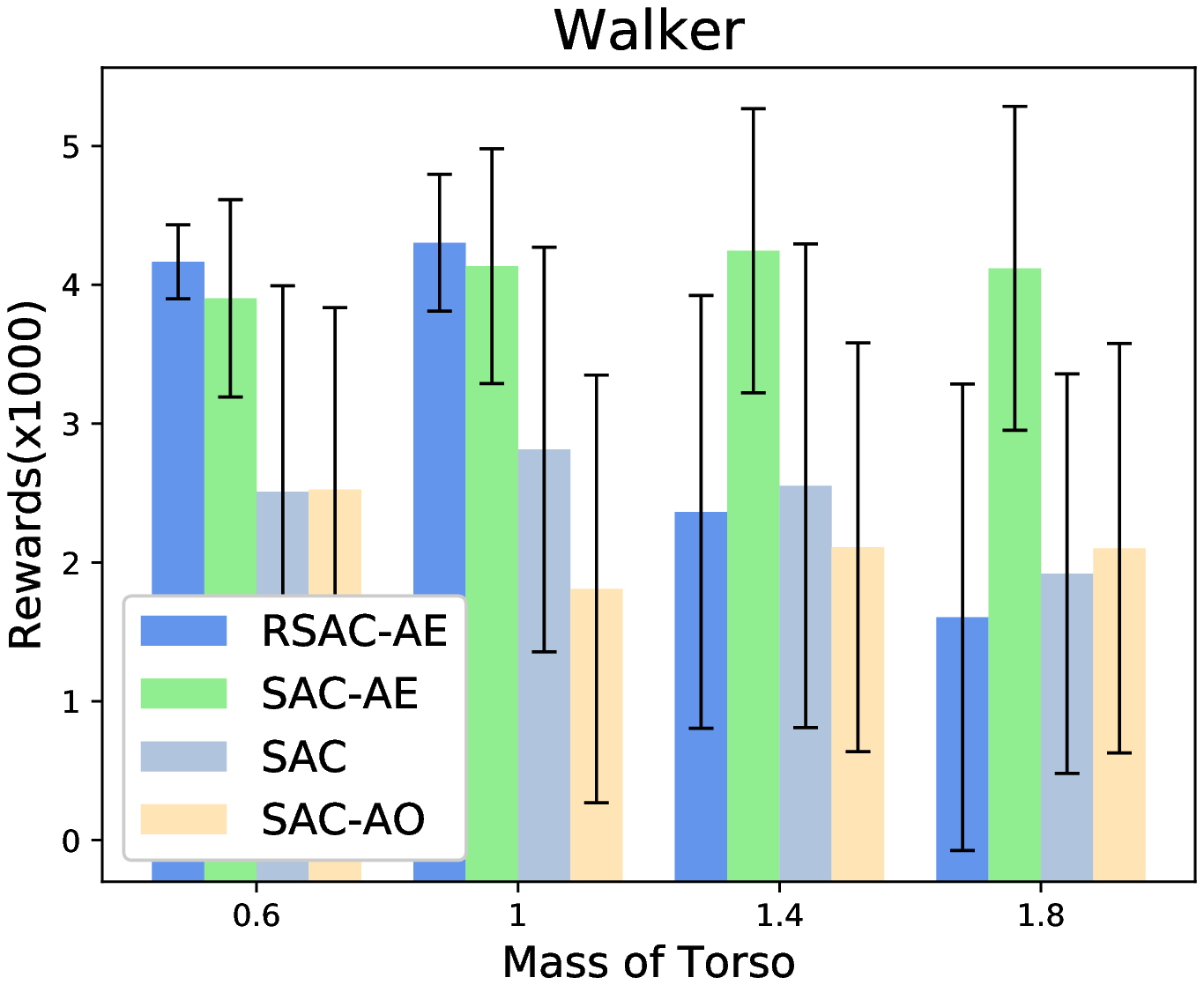}
\end{minipage}
\begin{minipage}[t]{0.235\textwidth}
\centering
\includegraphics[width=1\textwidth,height=3.3cm]{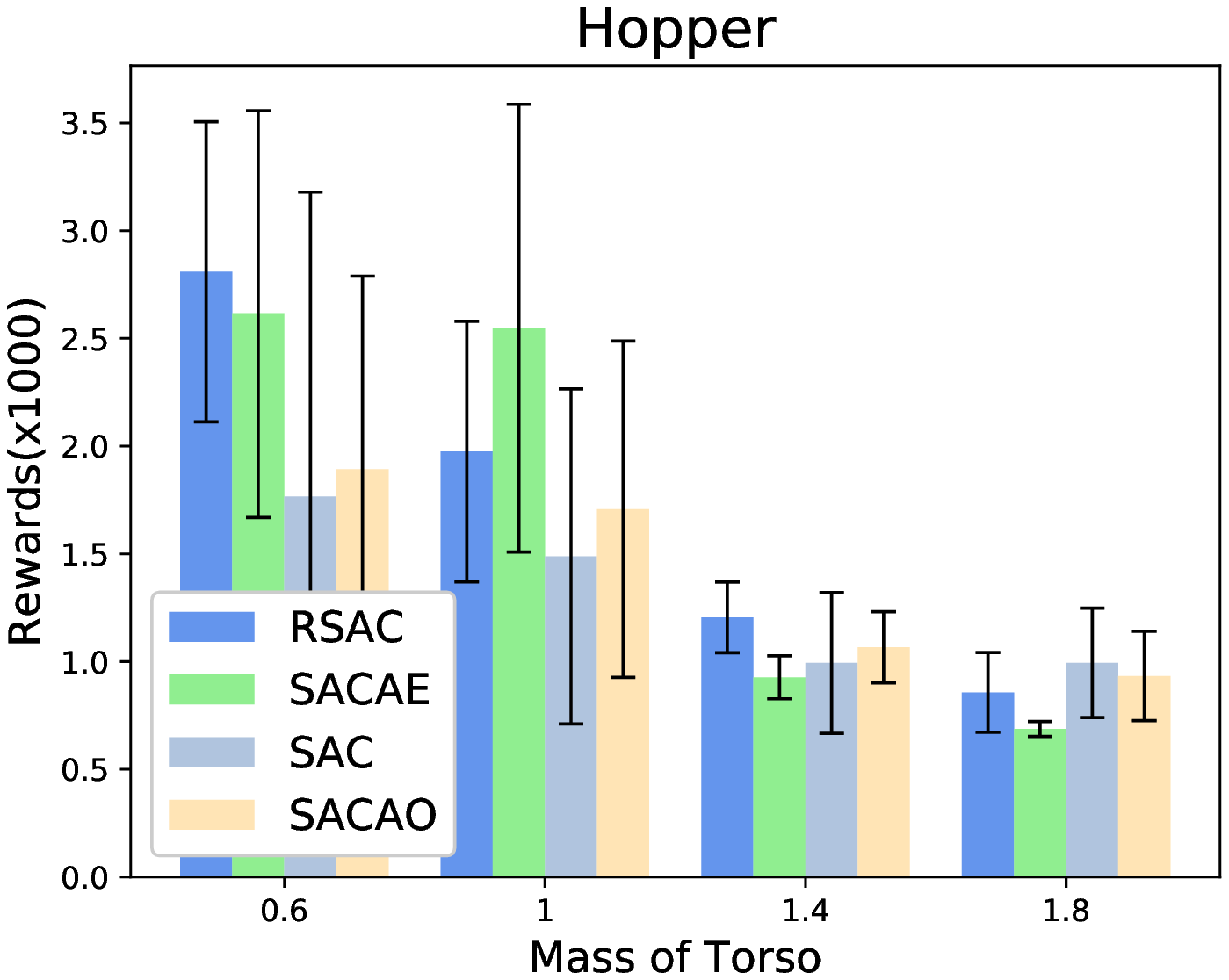}
\end{minipage}
\caption{The Main results. The attack parameter set $\Omega$ is a uniform distribution $U(\frac{1}{\kappa},\kappa)$ where $\kappa=1.5$ for all the environments.}
\label{MainResults}
\end{figure}

The final results are presented in Fig.~\ref{MainResults}.

As can be seen, RSAC-AE shows a leading performance in Pendulum, Cartpole and Hopper domains. In Walker domain, its performance degrades as the mass of torso increases. Nevertheless, SAC-AE outperforms other methods in this environment, proving the effectiveness of our adversarial training method. Consequently, we can conclude that (1) our attack algorithm helps improve the robustness of the agent (2) history trajectory and RNN help in handling with model misspecification (3) our robust training framework is better than an RMDP-based one.

\subsection{Investigative Experiments}
We have made several additional experiments to better explore the features of our proposed framework. All the experiments are made in Pendulum and Cart-Pole environments. 

\begin{figure}[htbp]
\centering
\begin{minipage}[t]{0.235\textwidth}
\centering
\includegraphics[width=1\textwidth,height=3cm]{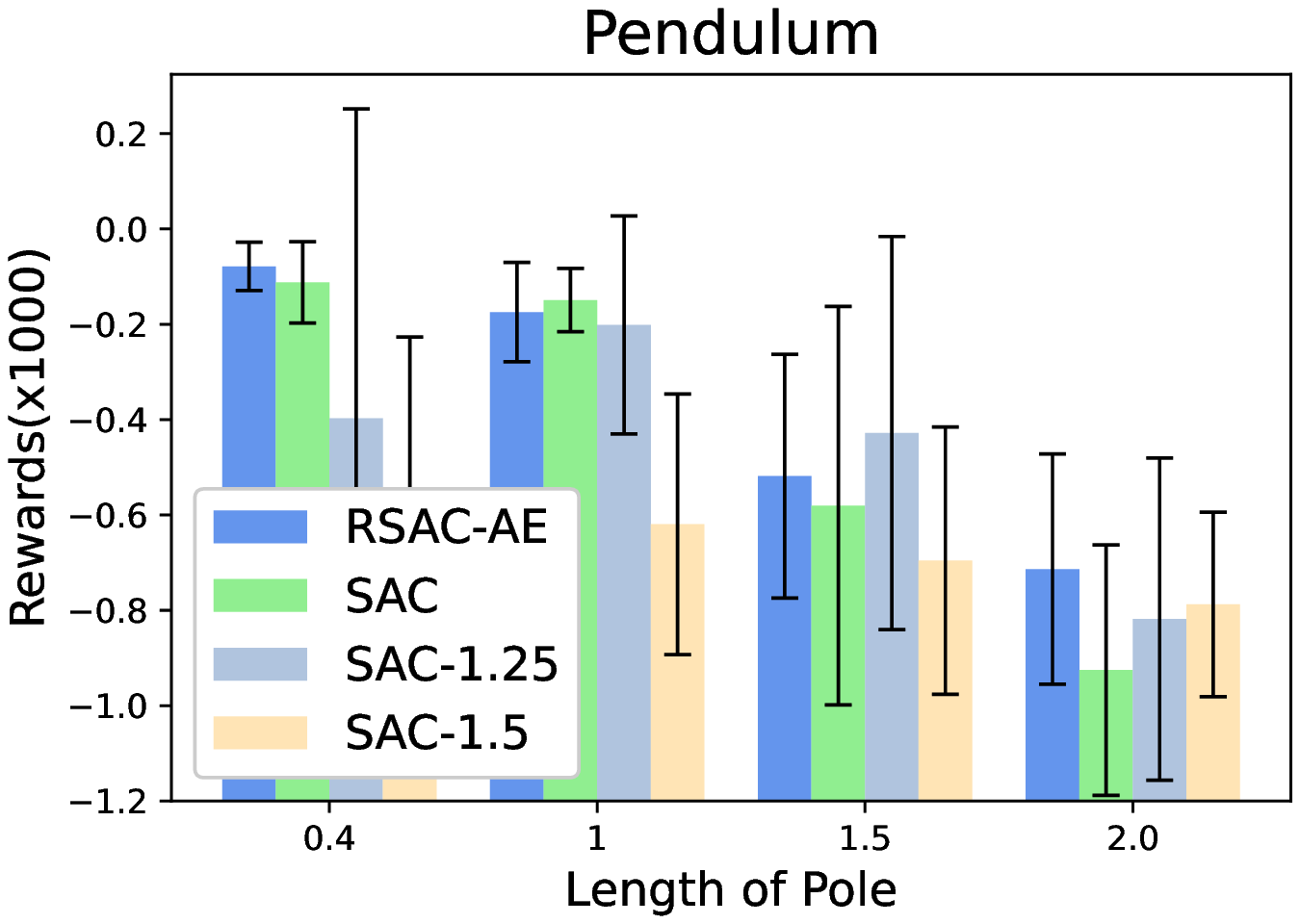}
\end{minipage}
\begin{minipage}[t]{0.235\textwidth}
\centering
\includegraphics[width=1\textwidth,height=3cm]{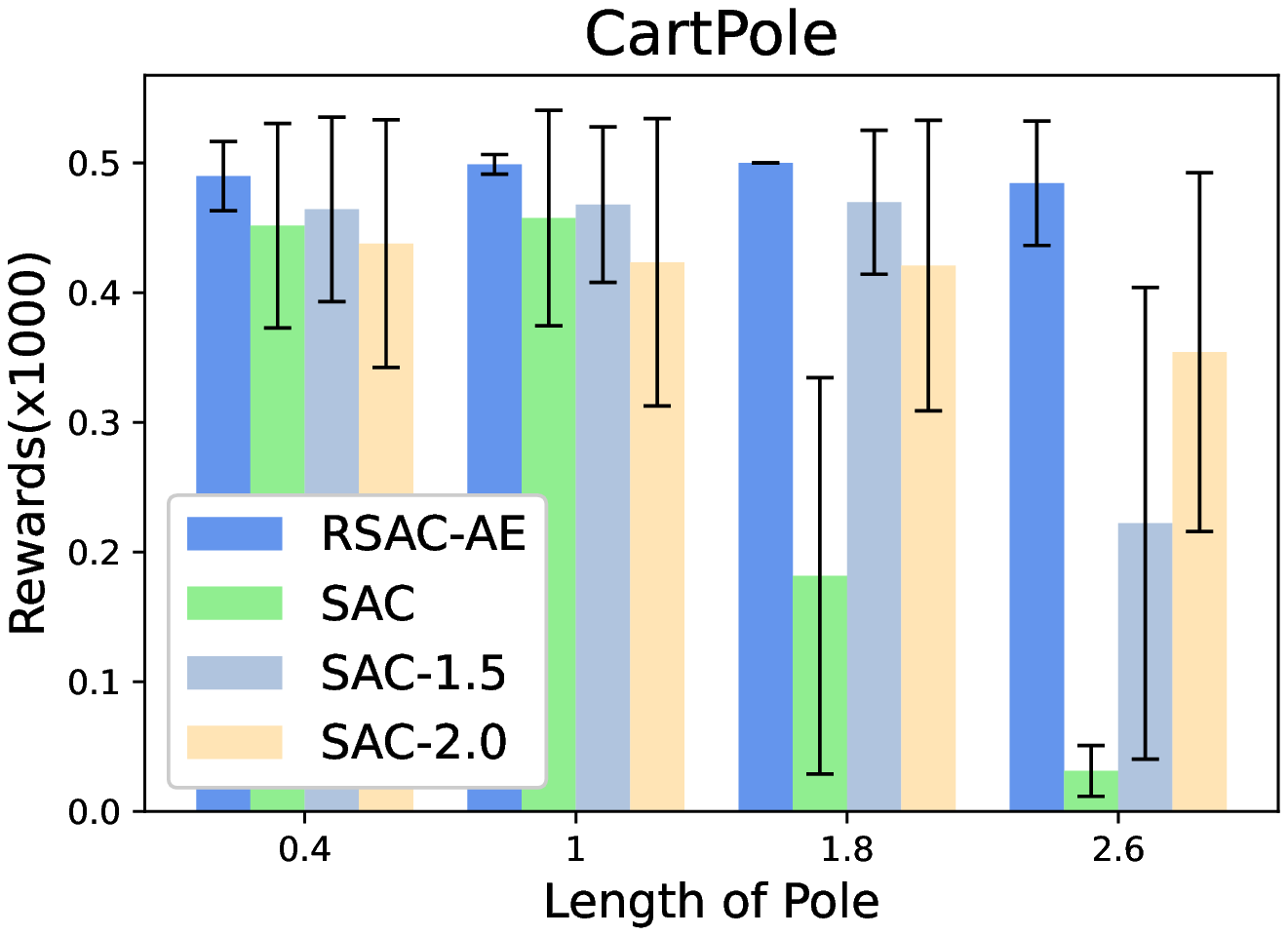}
\end{minipage}
\caption{Extensive Experiment: Modify the default environment of SAC}
\label{multiSAC}
\end{figure}

\textbf{Modify the default environment of SAC} In these four experiments, We notice a common phenomenon: the performance of the agent declines as $\omega$ increases. So the satisfying performance of RSAC may result from training in an environment with bigger $omega$ instead of our robust training framework. In this way, a baseline agent trained in an environment with a bigger $\omega$ should behave better. To test this hypothesis, we modified the default environment parameters as follows: for pendulum, the pole length is set to 1.25 and 1.5; for cart-pole, the pole length is 1.5 and 2.0. As comparison, the environment parameter set $\Omega$ for RSAC remains the same. As can be seen in Fig.~\ref{multiSAC}, the new models may transcend the original SAC model in environments with larger perturbation, but still cannot outperform RSAC. Another notable phenomenon is that training in an environment with too big $\omega$ can harm the overall performance. It may be too difficult for an agent to learn well in such difficult settings. 

\begin{figure}[htbp]
\centering
\begin{minipage}[t]{0.235\textwidth}
\centering
\includegraphics[width=1\textwidth,height=3cm]{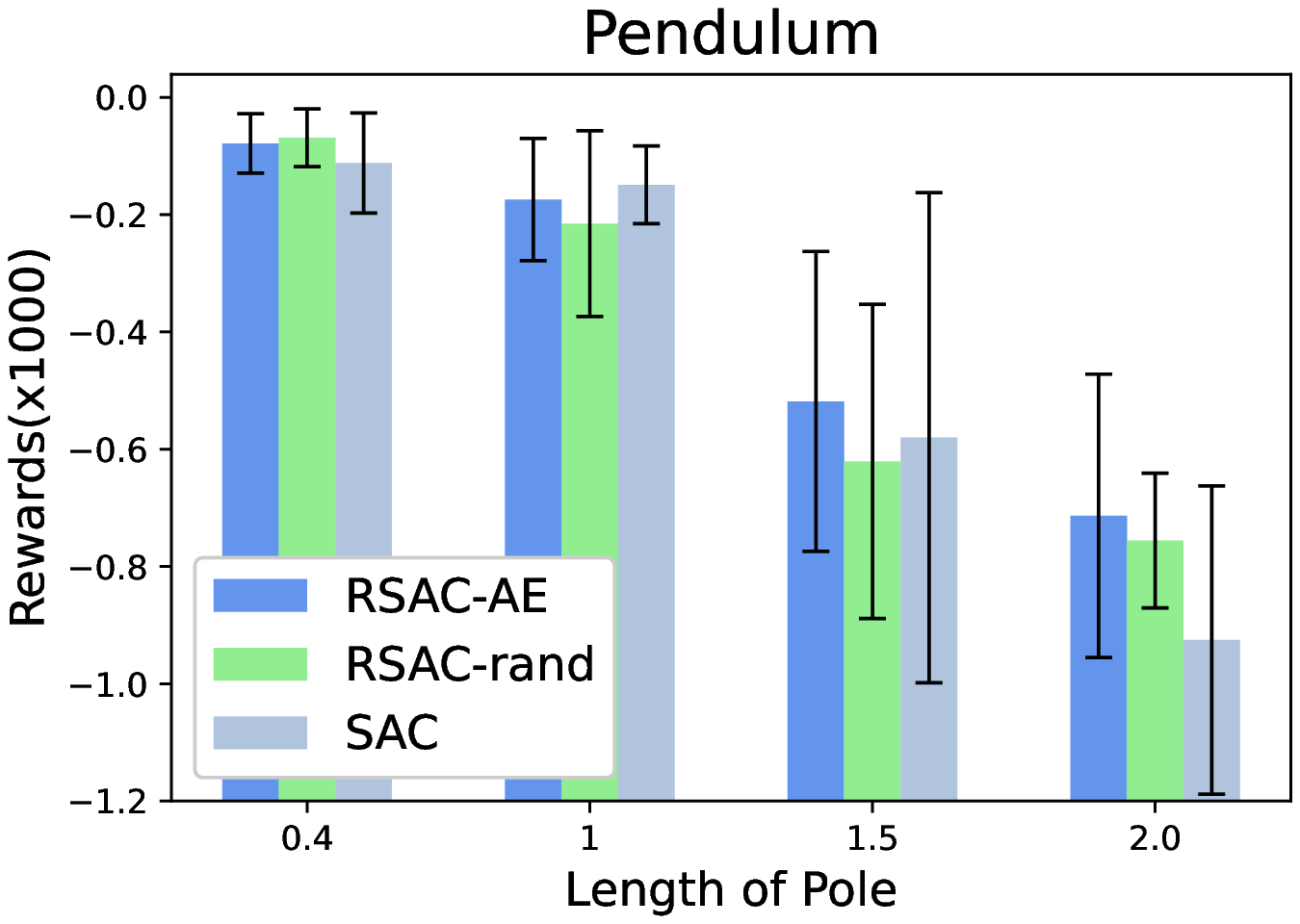}
\end{minipage}
\begin{minipage}[t]{0.235\textwidth}
\centering
\includegraphics[width=1\textwidth,height=3cm]{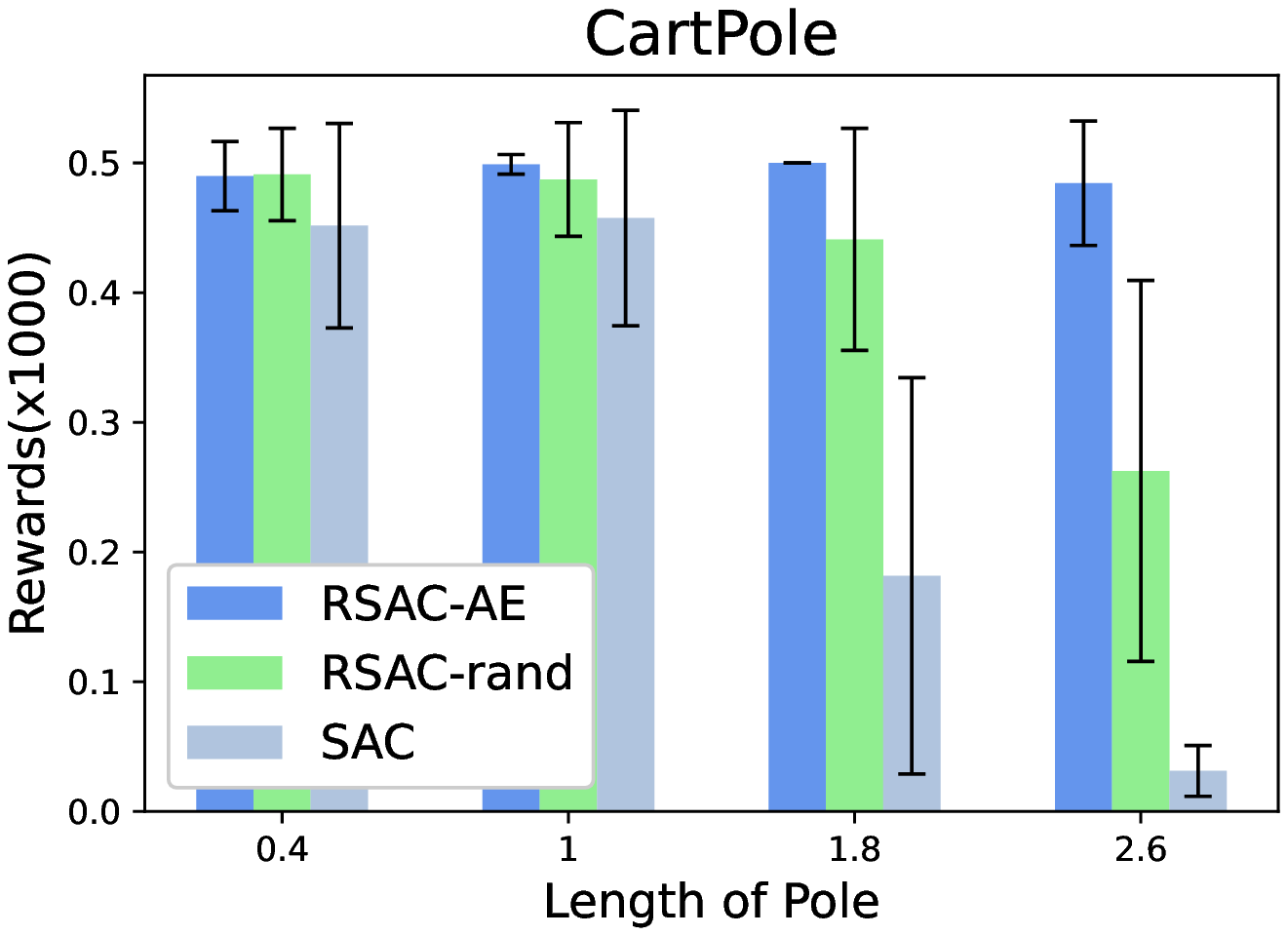}
\end{minipage}
\caption{Extensive Experiment: Train RSAC with random attack}
\label{randATK}
\end{figure}

\textbf{Train RSAC with random attack}
Our adversarial attack algorithm is preliminarily proved to be valid. To further test its utility, we train RSAC agents with random attack, i.e. $\omega$ will vary periodically and the new value is sampled from $\Omega$. As presented in Figure \ref{randATK}, random attack also improves the robustness, but is less effective than adversarial attack.

\begin{table}[t]
\caption{Hyperparameters for R-SAC}
\label{training parameters}
\begin{center}
\begin{tabular}{|c|c|}

\hline 
\bf Hyperparameters & \bf Values \\
\hline 
optimizer         & Adam \cite{kingma2014adam} \\
learning rate         & 0.0003 \\
train frequency         & 1 \\
batch size         & 64 \\
layer norm         & None \\
activation function         & relu \\
 target update interval         & 1 \\
 gradient steps         & 1 \\
 memory length         & 10 \\
 hidden units         & FC1:64 GRU:64 FC2:8 FC3:64 FC4:64 \\
$\gamma$         & 0.99 \\
$\tau$         & 0.005 \\
\hline 
 
\end{tabular}
\end{center}
\end{table}

\subsection{Hyperparameters}
The shared training hyperparameters for R-SAC are given in Table~\ref{training parameters}. Hyperparameters for SAC are the same except that its neural networks have two hidden layers with 64 units each. Training steps for the four domains are 60k, 100k, 2M and 2M. Note agents based on SAC-AO are trained without noise in the first 50$\%$ steps. After that, adversarial training begins. Buffer size is 50k for Pendulum $\&$ CartPole and 500k for the other two.   

Attack parameters for env-based-attack (Algorithm~\ref{alg:atkenv}) are list as follows: $t_{th}=40$, $n=20$, $\Omega=[0.67,1.5]$. $\sigma_{th}$ for Pendulum, CartPole, Walker and Hopper are 0.2, 0.2, 0.9 and 0.9. The hyperparameters $\beta$ in Algorithm~\ref{SAC-AO} for these four domains are 0.1, 0.1, 0.02 and 0.008.  
 
\section{Conclusion}
In this paper, we put forward a framework for robust reinforcement learning against model misspecification, a situation where the environment dynamics are perturbed. We discover a drawback of the commonly-used Robust Markov Decision Process (RMDP) framework and prove treating environment parameters as a part of the state results in a better strategy. Since the dynamics are unknown to agents, we utilize Partial Observable Markov Decision Process (POMDP) modeling and history trajectory. As far as we know, we are the first to do so. Additionally, we come up with an novel adversarial attack method to assist robust training. Moreover, we extend an acclaimed RL algorithm, Soft Actor-Critic, to POMDP scenarios. We also design a specialised network architecture for this task. Our experiments in four gym domains confirm the effectiveness of history trajectory as well as the adversarial attack algorithm.  



\end{document}